\documentclass[sigconf]{aamas}
\usepackage{placeins}
\usepackage{xcolor}
\usepackage{subcaption}

\usepackage{balance} 
\usepackage[ruled,vlined]{algorithm2e}
\usepackage{xcolor}
\usepackage[normalem]{ulem}

\SetKwInOut{Parameter}{parameter}

\setcopyright{ifaamas}
\acmConference[AAMAS '21]{Proc.\@ of the 20th International Conference on Autonomous Agents and Multiagent Systems (AAMAS 2021)}{May 3--7, 2021}{Online}{U.~Endriss, A.~Now\'{e}, F.~Dignum, A.~Lomuscio (eds.)}
\copyrightyear{2021}
\acmYear{2021}
\acmDOI{}
\acmPrice{}
\acmISBN{}

\acmSubmissionID{356}

\title[AAMAS-2021 Formatting Instructions]{An Autonomous Negotiating Agent Framework with Reinforcement Learning Based Strategies and Adaptive Strategy Switching Mechanism}

\author{Ayan Sengupta}
\affiliation{
  \institution{NEC Corporation}
  \city{Tokyo, Japan}}
\email{a-sengupta@nec.com}

\author{Yasser Mohammad}
\affiliation{
  \institution{NEC Corporation}
  \city{Tokyo, Japan}}
\email{y.mohammad@nec.com}

\author{Shinji Nakadai}
\affiliation{
  \institution{NEC Corporation}
  \city{Tokyo, Japan}}
\email{nakadai@nec.com}

\begin{abstract}
	Despite abundant negotiation strategies in literature, the complexity of automated negotiation forbids a single strategy from being dominant against all others in different negotiation scenarios. To overcome this, one approach is to use mixture of experts, but at the same time one problem of this method is the selection of experts, as this approach is limited by the competency of the experts selected. Another problem with most negotiation strategies is their incapability of adapting to dynamic variation of the opponent's behaviour within a single negotiation session resulting in poor performance. This work focuses on both, solving the problem of expert selection and adapting to the opponent's behaviour with our Autonomous Negotiating Agent Framework. This framework allows real-time classification of opponent's behaviour and provides a mechanism to select, switch or combine strategies within a single negotiation session. Additionally, our framework has a reviewer component which enables self-enhancement capability by deciding to include new strategies or replace old ones with better strategies periodically. We demonstrate an instance of our framework by implementing maximum entropy reinforcement learning based strategies with a deep learning based opponent classifier. Finally, we evaluate the performance of our agent against state-of-the-art negotiators under varied negotiation scenarios.
	
\end{abstract}


\newcommand{\BibTeX}{\rm B\kern-.05em{\sc i\kern-.025em b}\kern-.08em\TeX}

\begin{document}


\pagestyle{fancy}
\fancyhead{}


\maketitle


\section{Introduction}
Negotiation has been studied for a long time from different perspectives like game theory~\cite{raiffa2002negotiation}, business~\cite{bazerman1992reversals}, psychology~\cite{adams1965inequity}, neuroeconomics~\cite{coricelli2009neural} and many more. With the progress of AI-technologies, automated negotiation allows collaboration and negotiation among AI-enabled parties.
Automated negotiation aims to achieve win-win deals for all parties, while simultaneously reducing the time and effort, thus adding significant value to society as a whole~\cite{lin2008negotiating}. But the complexity of automated negotiation still hinders the deployment of autonomous agents in real-world applications~\cite{baarslag2017will}.\par
Though much research already existed in developing negotiation strategies, Automated Negotiating Agents Competition (ANAC) brought significant improvements in strategy development~\cite{jonker2017automated}. In spite of such improvements in strategy design, there is no single strategy that is optimal for all possible domains~\cite{ilany2016algorithm}. One natural solution is to choose a pool of strategies and  use the approach of mixture of experts  during negotiation. At the same time, one needs to choose an appropriate initial set of expert strategies to excel. The questions that can originate while designing such an algorithm are these: What initial set of strategies should we select? On what conditions should we switch strategies? How to improve the initial set of chosen strategies? In this work we give a solution to all three of  these questions by introducing our autonomous negotiating agent framework.\par
 The contributions of this work to the existing research in this domain are three-fold. Firstly, we propose an autonomous negotiating agent framework, which facilitates the creation of autonomous negotiating agents capable of classifying opponent's behaviour and adaptively change strategies within a single negotiation session to reach better agreements. Secondly, we propose a mechanism to update the base strategies in an algorithmic manner to improve the overall performance. Finally, we validate this framework and provide insights in general about autonomous negotiating agents by evaluating it extensively against state of the art negotiators.\par
The rest of the paper is organized as follows: Section~\ref{sec:related_work} gives a sketch of related work in this domain, Section~\ref{sec:neg_setting} provides the introduction to negotiation settings. Section~\ref{sec:framework} gives a detailed description of each of the components in our framework and Section~\ref{sec:experiment} describes the experimental setup. Section~\ref{sec:result} shows the evaluations of our framework and finally, we conclude with Section~\ref{sec:conclusion} by discussing the limitations and provide direction for future research.
\section{Related Work}\label{sec:related_work}
A considerable amount of literature has already been published on autonomous negotiation strategies. However, in recent years, the success of reinforcement learning (RL) algorithms in different fields has drawn significant attention to its application in autonomous negotiation~\cite{tesauro2000pricing, bakker2019rlboa}. A part of our work falls under the above mentioned domain. Additionally, the other part of our work is at the intersection of the domains of opponent classification and strategy selection in autonomous negotiation. In this section we discuss the work done in both of these domains.

\subsection{RL in Autonomous Negotiation}
Previously many computational methods including Bayesian Learning~\cite{zeng1998bayesian, hindriks2008opponent} and Genetic Algorithm~\cite{matos1998determining, choudhary2019evolutionary, lau2006evolutionary}  have been used in automated negotiation for developing and evaluating negotiation strategies. Then again, in the last couple of decades several studies have looked at the application of reinforcement learning (RL) algorithms like Q-learning~\cite{chen2013bilateral, tesauro2000pricing, tesauro2002pricing, sridharan2002multi, bakker2019rlboa} and REINFORCE~\cite{sunder2018prosocial} in automated negotiation. Recently, Deep Reinforcement learning (DRL) has been used to learn the target utility values~\cite{bagga2020learnable}, the acceptance strategy~\cite{razeghi2020deep} or both bidding and acceptance strategies~\cite{chang2020multi}. Moreover, authors of~\cite{bagga2020} have also shown application of DRL in concurrent bilateral negotiation.\par
Bakker et al. introduced RLBOA framework~\cite{bakker2019rlboa} based on the BOA architecture~\cite{baarslag2014decoupling} for automated negotiating agents,
where they trained the  bidding strategy of the agent using Q-learning. Their approach involves discretizing utility space and using opponent modelling to choose next offer from a set of offers, where the set of offers at each time step depends on the action taken. A limitation of this method is the loss of information due to discretization of utility space and that leads to further dependence on opponent modelling for the choice of next offer. In contrast to their work, we do not use opponent modelling while training bidding strategies. Moreover, we train the bidding strategy using DRL on continuous state and action spaces.\par
The authors of~\cite{chang2020multi} have used DRL algorithms for training both bidding and acceptance strategies in continuous state and action spaces. 
The state space and the action space for their approach includes actual offer from the outcome space and hence limits the scope to a particular negotiation scenario.
Furthermore, in their approach one needs to train both acceptance and bidding strategy for every domain. 
Moreover, experimental setup and evaluations were done against fixed preference profiles, which limits the scope of applicability. In contrast, our approach considers the utility value of the offers projected to self utility axis, thus making our bidding strategies applicable to multiple negotiation scenarios. Additionally, we show the generality of our approach by evaluating in varied negotiation scenarios while training in a single negotiation domain.\par
Furthermore, evaluations in both~\cite{bakker2019rlboa} and~\cite{chang2020multi} are against primitive agents only, whereas we evaluated our approach against GENIUS~\cite{Genius} based ANAC~\cite{jonker2017automated} winning agents.

\subsection{Opponent Modelling and Strategy Selection in Automated Negotiation}
The Opponent modelling is a fundamental component of BOA architecture proposed in~\cite{baarslag2014decoupling}. Commonly, opponent models attempt to learn one or more of the following opponent's attribute: acceptance strategy, deadline, utility function or bidding strategy~\cite{baarslag2016learning}. However, our approach does not fit any of these usual types of opponent models. Unlike popular approaches of learning the bidding strategy, we classify an opponent depending on the history of bids. In fact our problem of classifying the opponent falls under the domain of continuous opponent strategy classification. Under this domain for instance, authors of~\cite{schadd2007opponent} used a hierarchical approach with fuzzy models to perform the opponent strategy classification in a real time strategy game.
Preference profile learning by classifying the negotiation trace   was done in \cite{lin2006bounded, lin2008negotiating, hindriks2008opponent} using Bayesian learning to determine the best match for the opponent's preference profile.
However, in this work we classify the opponents bidding behaviour periodically with respect to a set of  negotiator's  bidding behaviours and select an appropriate strategy for negotiation within a single negotiation session.
\par
Inspired from algorithm selection method~\cite{leyton2003portfolio}, authors of~\cite{ilany2016algorithm} developed
a meta-agent, which predicts the performance of a set of bilateral negotiation negotiators based on features of domains, and accordingly chooses the negotiator expected to
perform best for the given negotiation scenario. Extending this idea to multilateral negotiation settings and using the approach of mixture of experts, authors of~\cite{gunecs2017collective} solved the problem of how to combine multiple experts. However, in all these approaches a single negotiator is selected throughout a negotiation session. In contrast to that, this paper focuses on selection and switching (or combination) of strategies within a single negotiation session based on the opponent behaviour.


\section{Negotiation Settings}\label{sec:neg_setting}
A bilateral automated negotiation  is a negotiation between two automated entities. We will denote these entities as \emph{negotiators}. A negotiation setting consists of a negotiation protocol, the concerned negotiators and negotiation scenario~\cite{baarslag2014bid}. A negotiation protocol defines the rules of the encounter, specifying which actions each negotiator can perform at any given moment. A negotiation scenario consists of the preference profiles of each negotiator and the negotiation domain. In this work, a \emph{strategy} of a negotiator is the combination of an acceptance strategy and a bidding strategy. Additionally we denote the opponent negotiators drawn from GENIUS platform~\cite{Genius} as \emph{agents}. 
\par

\par
The negotiation protocol used throughout this paper is the stacked alternating offers
protocol. Under this protocol, a negotiation session consists of rounds of consecutive
turns where each negotiator can either make an offer, accept offer, or
walk away from the negotiation~\cite{aydougan2017alternating}.
The negotiation session ends if  both negotiators find a joint
agreement or a deadline is reached or one of the negotiators decides to walk away from the negotiation resulting in no agreement. The deadline can be measured in
number of rounds or actual wall-time. Negotiations are non-repeated, that is
one negotiation session cannot impact actions of any negotiator in subsequent sessions.

\par
A negotiation domain consists of one or more issues. To reach an agreement, the negotiators must settle on a specific value for each negotiated issue. The outcome space of a negotiation domain denoted by $\Omega$  is the set of all possible negotiation outcomes. The outcome space can be defined as the Cartesian product of negotiation issues
 and is formally denoted as $\Omega = \{ \omega_1, \cdots, \omega_n \}$ where $\omega_i$
is a
possible outcome and $n$ is the carnality of outcome space.
A preference profile or utility profile defines a preference
order $\leq$ that ranks the outcomes in the outcome space. Usually, a preference profile of a negotiator is specified by a utility function, which assigns a utility value to an outcome $\omega_i$ denoted by $U(\omega_i)$.
Utility functions are private information and the negotiators only know their own utility functions. The preference profile of an agent  also specifies a reservation value. The reservation value $u_r$ is the utility that the negotiator receives in case of no agreement.

\section{Proposed Negotiator Framework}\label{sec:framework}

\begin{figure}[ht]
  \centering
  \includegraphics[width=0.46\textwidth]{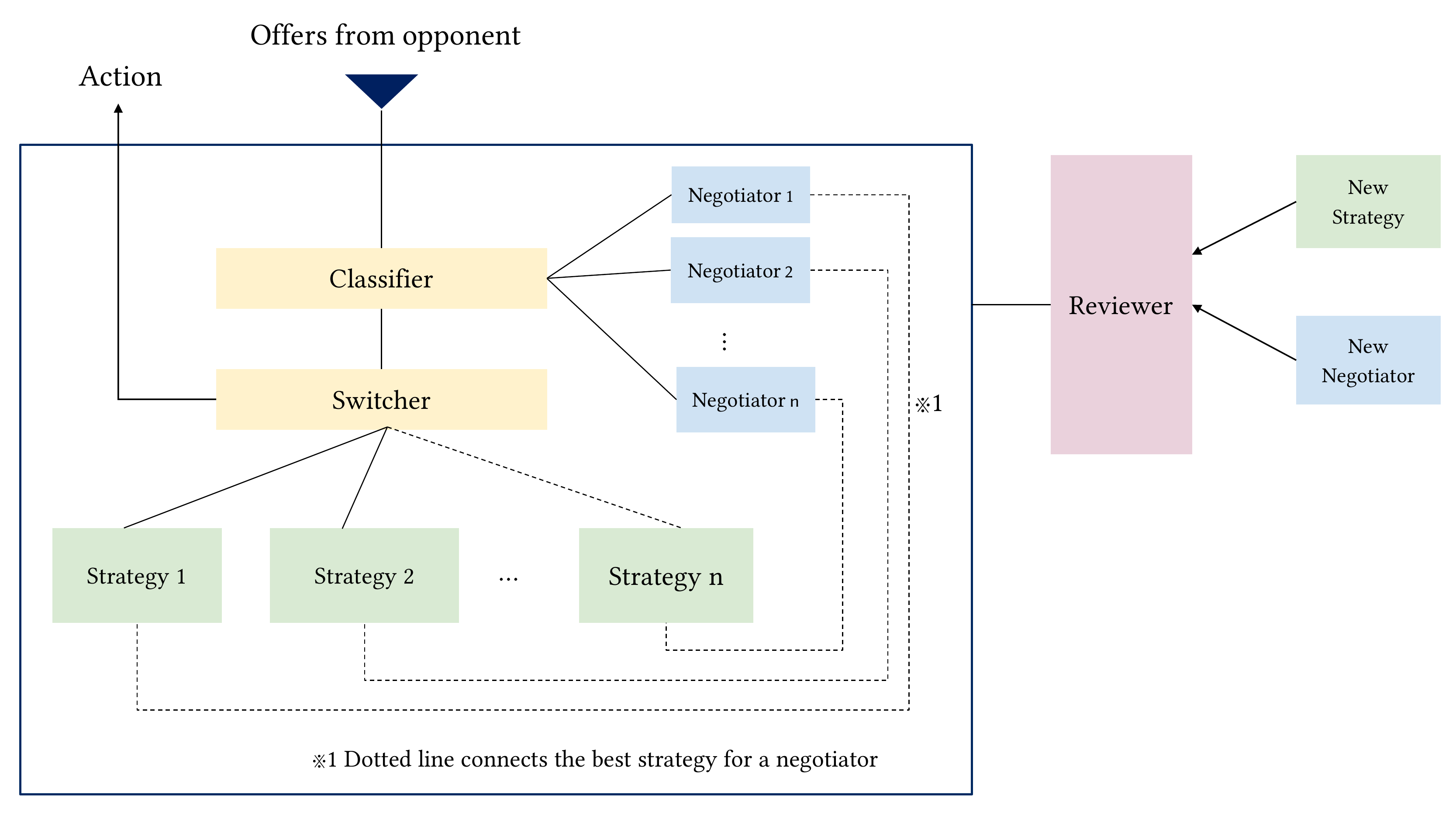}
  \caption{Block diagram of proposed framework showing the following blocks: $n$ base negotiators blocks (blue), $n$ trained strategies block (green), the classifier block (yellow), strategy switching block (yellow) and the reviewer block (purple). The dashed lines connecting a single negotiator to a single strategy represents  negotiator-strategy pair. The components inside the solid box are utilised within a negotiation session whereas the blocks outside the solid box are used outside a negotiation session.}
  \label{fig:framework_block}
  \Description{Classifier model having three consecutive 1D-CNN layers followed by two Dense layer for $k=20$ and $N=7$}
\end{figure}

In this section we provide the structure and explain the details of our proposed Autonomous Negotiating Agent Framework, a framework that facilitates the creation of autonomous negotiating agents which are capable of classifying opponents in real time and switching strategies accordingly within a single negotiation session.  First, we introduce the components of our framework and then describe an approach for designing each of the components. The proposed framework is comprised of four main components: negotiator-strategy pair, opponent classifier, strategy switching mechanism and reviewer. Figure~\ref{fig:framework_block} outlines all the components of the framework, while each of them are discussed in the remainder of the section.\par

The first component of our framework is a set of negotiator-strategy pairs, where negotiators can be any autonomous negotiating agents and strategies are bidding strategy trained against the negotiators in addition to fixed acceptance strategy for each chosen negotiator. The framework gives the user the flexibility to choose any set of negotiators which we will call the base negotiators for the rest of this paper. 
To illustrate, the base negotiators can include from simple time or behaviour dependent strategies~\cite{faratin1998negotiation} to state-of-the-art negotiators like ANAC winning agents. 
In this work, as described in Section~\ref{Framework:DRL}, we have trained deep reinforcement learning (DRL) based bidding strategies against each negotiator to form  \emph{negotiator-strategy} pairs and subsequently showed the superiority and generality for such class of strategies.\par
The second component is an opponent classifier that classifies the opponent's bidding behaviour with respect to the bidding behaviour of base negotiators. After every negotiation round during a negotiation session, the classifier takes a sequence of opponent's bid as input and 
accordingly assigns an estimated probability to each base negotiator.  In this work, we have used only the sequence of opponent's offers projected on the self utility axis as input since our implementation showed similar accuracy with additional information of self offers. Deep learning based approach, described in Section~\ref{Framework:Classfier} is used for training the opponent classifier and the results in Section~\ref{sec:result} show the versatility of such classifier in different negotiation scenarios.\par 

The third component is a switching mechanism, that switches or combines the strategies learned against base negotiators depending on the output of the opponent classifier. 
While the opponent classifier classifies the behaviour of the opponent negotiator at every time step, the switching mechanism has added flexibility of changing decisions after certain intervals. Note that the strategy switching as narrated in Section~\ref{framework:switcher} is performed within a single negotiation session, and hence makes our negotiator framework adaptive.\par

All the three aforementioned components are active components that function when a negotiation is underway. Whereas, the Reviewer is a passive component that does not actively take part in the negotiation process. Outside the negotiation session, the Reviewer provides a mechanism that decides if a new negotiator or a new strategy should be included in the framework. To the best of our knowledge, all the meta-agent strategies proposed in literature do not have a mechanism that can enhance their capability by evaluating and adding new strategies. This component is crucial to the design of our framework as it insures the framework against depreciation in the future.
In Section~\ref{framework:reviewer}, we provide the algorithm of the Reviewer and in the remaining section, we discuss the approaches for building each component of our framework.

\subsection{Deep Reinforcement Learning Based Strategies}\label{Framework:DRL}
The prime components of our framework are the negotiation strategies trained against the base negotiators. The whole framework is based on the presumption that one can successfully learn an effective strategy against each of the base negotiators. Additional requirement is that the approach should facilitate the framework to perform well in a domain-independent manner.
Due to recent success of RL algorithms in training strategies of automated negotiators, we used Soft Actor-Critic~\cite{haarnoja2018soft, haarnoja2018soft2}, a DRL algorithm to train a bidding strategy against each base negotiator. For acceptance strategy we adopted the approach of combined acceptance condition as proposed in~\cite{baarslag2014effective}. In contrast to RLBOA framework in~\cite{bakker2019rlboa}, no opponent model component is used while training against an opponent negotiator.

\subsubsection{State Space, Action Space and Reward}
A major problem for developing a domain independent negotiator framework is the fact that the outcome space $\Omega$ varies significantly across different negotiation scenarios. Moreover the offers $\omega_i \in \Omega$ are usually non numerical in nature, which again demands an approach to convert the offers to numerical value in a meaningful way. To overcome both of these problems, we took a similar approach to~\cite{bakker2019rlboa} and represented every outcome $\omega_i$ by $U_s(\omega_i)$ where $U_s$ is the self utility function. To avoid any information loss we considered the continuous outcome space rather than discretizing it. For our DRL approach everything in the negotiation scenario including the opponent is considered as the environment. Let us denote the state and action in an environment as $s_t$ and $a_t$ respectively. The state consists of only the information about the offers and the action determines what utility value to bid next. For a negotiation session with time limit $T$, we defined our state space and action space as
\begin{eqnarray*}
s_t =&\{ t_r, U_s(\omega_{s}^{t-2}), U_s(\omega_{o}^{t-2}), U_s(\omega_{s}^{t-1}),\\ &U_s(\omega_{o}^{t-1}), U_s(\omega_{s}^{t}), U_s(\omega_{o}^{t})\} \\
a_t =& u_s^{t+1} \hspace{0.8em} \text{such that}\hspace{0.8em} u_r < u_s \leq 1
\end{eqnarray*}
where $t_r$ denotes the relative time, $\omega_s^t$ and $\omega_o^t$ denotes offers by self and opponent at time step $t < T$ respectively. The self reservation value is denoted by $u_r$ and $u_s^{t+1}$ denotes the utility value of the next offer. To get the actual offer from the utility value we need an inverse map $U_s^{-1}:u_s\rightarrow \omega_s$ of the self utility function $U_s$
which can be a one-to-one or one-to-many mapping. One simple way of defining the inverse utility function is given in Equation~(\ref{equation:inverse_utility}).
\begin{equation} \label{equation:inverse_utility}
\begin{split}
U_s^{-1}(u_s) &= \operatorname*{argmin}_\omega f(\omega),  \hspace{0.8em} \text{where}\\
f(\omega) &= (U_s(\omega) - u_s)^2 \hspace{0.8em} \forall \hspace{0.8em} \omega \in \Omega.
\end{split}
\end{equation}
The goal of the strategies trained is to maximise the average utility against the corresponding base negotiator. So the reward function $R$ is defined as
\begin{eqnarray*}
R(s_t, a_t, s_{t+1}) =
\begin{cases}
     U_s(\omega_{a}),& \text{if there is an agreement } \omega_a\\
    -1,              & \text{for no agreement and } s_{t+1}\\ & \text{is terminal state,}\\
    0, & \text{otherwise.}
    \end{cases}
\end{eqnarray*}
There is an immediate reward of $0$ after every step in a negotiation session when negotiation has not ended.
\subsubsection{Soft Actor-Critic Algorithm}
Soft actor-critic (SAC)~\cite{haarnoja2018soft, haarnoja2018soft2} is an off-policy algorithm based on maximum entropy reinforcement learning that aims to  maximize both the expected reward and the policy's entropy. Policies with higher entropy have more randomness, which means that maximum entropy reinforcement learning learns a policy that has maximum randomness yet achieves a high reward. Normal reinforcement learning algorithms try to maximize the expected reward only. On the contrast the reason for maximizing the the entropy of the policy is to improve both the algorithm's robustness to hyperparameters and its sample efficiency~\cite{haarnoja2018soft}.In automated negotiation, this randomness is desirable to reduce the opponent's ability to predict the behaviour of an agent and exploit this information.\par



An optimal policy $\pi^*$ in entropy-regularized reinforcement learning can be expressed as
\begin{eqnarray*}
\pi^* = \arg \max_{\pi} \mathbb{E}_{\pi} \left[{ \sum_{t=0}^{\infty} \gamma^t \bigg( R(s_t, a_t, s_{t+1}) + \alpha H\left(\pi(\cdot|s_t)\right) \bigg)}\right],
\end{eqnarray*}
where $R$ is the reward function, $\gamma$ is the discount factor, $H$ denotes the entropy of policy $\pi$ and $\alpha > 0$ is the entropy regularization coefficient. $s_t$ and $a_t$ denotes the state and action at time-step $t$ respectively.
Now, the corresponding action-value function $Q^{\pi}(s,a)$ for state $s$ and action $a$ can be expressed as 
\begin{eqnarray*}
Q^{\pi}(s,a) =  \mathbb{E}_{\pi} \left[{ \left. \sum_{t=0}^{\infty} \gamma^t  R(s_t, a_t, s_{t+1}) + \alpha \sum_{t=1}^{\infty} \gamma^t H\left(\pi(\cdot|s_t)\right)\right| s, a}\right]
\end{eqnarray*}
SAC concurrently learns a policy $\pi$ and two Q-functions. In our implementation we used the approach proposed by Haarnoja et al.~\cite{haarnoja2018soft2} where entropy regularization parameter $\alpha$ is also a trainable parameter. The details of the hyperparameters for our implemented model are provided in the supplementary material.
\subsection{Deep Learning Based Opponent Classifier} \label{Framework:Classfier}
Opponent modelling is a fundamental block of BOA architecture proposed in~\cite{baarslag2014decoupling}. 
Although we do not have an opponent modelling block while training the DRL based strategies, our framework contains a classifier that classifies an unknown opponent's bidding behaviour with respect to the base negotiator's behaviour in the framework.
Our approach uses 1D-Convolutional Neural Networks (1D-CNN) based classifier to classify an unknown opponent at every time step of a negotiation. In the following sections we will describe our classifier's input/output and the model architecture.
\subsubsection{Input and Output}
The input to the classifier is a sequence of offers by the opponent projected to self utility axis.
Similar to Section~\ref{Framework:DRL}, we will denote each offer as $\omega_i$ and the self utility value for the offer as $U_s(\omega_i)$. The choice of using self utility value ensures the input sequence to be numerical all the time and can be directly provided to the classifier without any requirement of pre-processing. Another significant benefit is, it allows the framework to work across different negotiation scenarios without retraining the classifier for each domain.
The output of the classifier are the estimated probabilities for each base negotiator. Let us denote the input to the classifier at the current negotiation time step $t$ as $\mathcal{I}^t_c$ and output as $\mathcal{O}^t_c$. Then,
\begin{eqnarray*}
\mathcal{I}^t_c &=& \{U_s(\omega_i)\}_{i={t-k}}^{i={t-1}} \hspace{1em}\text{where} \hspace{1em} k \in \mathbb{Z}_+ \hspace{0.8em} \text{and} \hspace{0.8em}k > 1. \\
\mathcal{O}^t_c &=& [p_1, \cdots , p_n]
\end{eqnarray*}
where $n$ is the number of base negotiators in the framework. $\mathbb{Z}_+$ denotes the set of positive integers. $p_i$ denotes the estimated probability of the opponent behaving as the $i^{th}$ negotiator. Values of input array $\mathcal{I}^t_c$ are zero before the first opponent offer, that is, $U_s(\omega_i)=0$ if $i<0$. $k$ is the window length and is fixed before training. Moreover, greater the value of $k$, greater is the information provided to the classifier.
\subsubsection{Classifier Model}
The input to the classifier at every time step $\mathcal{I}^t_c$ is a time-series. For such time series data, Long short-term memory (LSTM) or recurrent neural network (RNN) architectures perform incredibly well. But, a major problem with LSTM and RNN is that they require datasets of massive sizes and large computational resources for training. To overcome such difficulties 1D-CNNs have shown great promise~\cite{kiranyaz20191d}. 
1D-CNN based classifiers have been successfully used in structural damage detection~\cite{abdeljaber2017real,avci2020convolutional}, fault detection in modular multilevel converters~\cite{kiranyaz2018real}, condition monitoring in rotating mechanical machine parts~\cite{ince2016real, eren2019generic}. 

\par
 In our classifier model, we have consecutive 1D-CNN layers followed by consecutive Dense layers. The depth of the model will increase or decrease with the increase or decrease of the window length $k$ respectively. Moreover, a greater value of $k$ will result in a larger part of history of opponent offers to be considered by the classifier.  This will reduce the model accuracy. On the other hand a very small value of $k$ will make the model myopic and will be error prone in classification. Some hyperparameter tuning is required for adjusting the value of the window length. The model architecture and the hyperparameters are provided in the supplementary material.


\subsection{Strategy Switching Mechanism} \label{framework:switcher}
Depending on the output  of the classifier $\mathcal{O}^t_c$, this component switches or combines the strategies to take the next action. The algorithm for the switching mechanism is provided in Algorithm~\ref{algorithm:switcher}. Although the opponent classification is done at every time step, the approach of choosing next offer need not change after every time step. 
In Algorithm~\ref{algorithm:switcher}  the parameter $\beta_i$ where $i\in[1,n]$ tunes the algorithm from a hard switcher to a combination mechanism.  The algorithm becomes a pure switching algorithm if $\beta_i=1$ and $\beta_k = 0 \;\; \forall \;\; k\neq i.$

\SetCommentSty{\triangleright}

\begin{algorithm}
\DontPrintSemicolon
\SetKwInput{KwInput}{Input}                
\SetKwInput{KwOutput}{Output}
\SetKwData{Data}{Data}
\SetAlgoLined
\KwInput{$\mathcal{O}^t_c = \{p_1,\cdots,p_n \}$ from the opponent classifier}
\KwData{$\mathcal{S}_b =\{s_1,\cdots,s_n \},$ the set of base strategies corresponding to the set of base negotiators $\mathcal{N}_b=\{N_1,\cdots,N_n \}$}
\KwOutput{Action: next offer or Accept }

Choose initial strategy $s_{init} \in S_b$, $S_{action} = s_{init}$\;

  $i = argmax(\mathcal{O}^t_c)$ where $N_i$ denotes the base negotiator with highest classification probability\;
  \eIf{action by strategy $s_i$ is Accept}{
   Accept opponent's offer\;
   }
    {
   $w^{t+1} = U_s^{-1}(\sum_{k=0}^{k=n} \left\{ \beta_k*u_{s_k})\right\}$ where $u_{s_k}$ is the utility value by strategy $s_k$ for next time step and $\beta_k$ is the weight parameter\;
   }
   
 \caption{Algorithm for strategy switching} \label{algorithm:switcher}
\end{algorithm}

\subsection{Reviewer Mechanism} \label{framework:reviewer}

This component enables the addition of new negotiators or strategies or both to the RL-agent instantiated by our framework. To show the basic operation of this component, we implemented an evaluation based approach for the Reviewer. The algorithm is provided in Algorithm~\ref{algorithm:reviewer} where the parameters $\alpha$ and $\beta$ are threshold parameters. When a new negotiator  $\mathcal{N}_{new}$ is introduced to the  Reviewer, first a new strategy $\mathcal{S}_{train}$ is trained against it. Subsequently, $\mathcal{S}_{train}$ and the  RL-agent are evaluated against $\mathcal{N}_{new}$. Finally, depending of the parameter $\alpha$ if the evaluation with the $\mathcal{S}_{train}$ is better in comparison with RL-agent, then the reviewer will provide confirmation and $\mathcal{S}_{train}$ will added to the pool of strategies and the classifier will be retrained with a new class  $\mathcal{N}_{new}$.

\begin{algorithm}[!ht]
\DontPrintSemicolon
\SetKwInput{KwInput}{Input}                
\SetKwInput{KwOutput}{Output}
\SetKwData{Data}{Data}
\SetKwInOut{Parameter}{Parameter}
\SetKwFor{For}{for (}{) }
\SetAlgoLined
\KwInput{New strategy $\mathcal{S}_{new}$ or new negotiator $\mathcal{N}_{new}$}
\KwData{$\mathcal{S}_b =\{s_1,\cdots,s_n \},$ the set of base strategies corresponding to the set of base negotiators $\mathcal{N}_b=\{N_1,\cdots,N_n \}$, \emph{Eval} function that provides an evaluation score of strategy for a negotiator.}
\SetAlgoNoLine%
\SetKwFunction{FMain}{StrategyEvaluation}
  \SetKwProg{Pn}{Function}{:}{}
  \Pn{\FMain{$\mathcal{S}_{test}$, $N_i$, $s_i$, $\beta$}}{
    $\hat{e}_k=Eval(N_i, \mathcal{S}_{test})$\;
  $e_k=Eval(N_i, s_i)$\;\SetAlgoLined
  \eIf{$\hat{e}_k \geq \beta*e_k$}
  {\KwRet Accept and replace $s_i$ with $\mathcal{S}_{test}$\;}
    {\KwRet Reject\;
  }}\SetAlgoLined
  \If{ Input is $\mathcal{N}_{new}$}{
   Train new strategy $\mathcal{S}_{train}$ against $\mathcal{N}_{new}$\;
   $e_f = Eval(\mathcal{N}_{new}, \text{ RL-agent})$\;
   $e_s = Eval(\mathcal{N}_{new}, \mathcal{S}_{train})$\;
   \eIf{$e_s \geq \alpha*e_{f}$}{
   Accept $\mathcal{N}_{new}$ and $\mathcal{S}_{train}$\;}{Reject}
   \For{$k \in [1,n]$}{
  StrategyEvaluation($\mathcal{S}_{train}$, $n_k$, $s_k$, $\beta$)}}
  \If{Input is $\mathcal{S}_{new}$}
  {\For{$k \in [1,n]$}{
  StrategyEvaluation($\mathcal{S}_{New}$, $n_k$, $s_k$, $\beta$)}}

 \caption{Algorithm for Reviewer component} \label{algorithm:reviewer}
\end{algorithm}

Moreover, The new trained strategy $\mathcal{S}_{train}$ is cross-evaluated with base negotiators and compared with the base strategies. Depending on the evaluation and parameter $\beta$, base strategies may be updated with $\mathcal{S}_{train}$. In this manner the Reviewer  provides a mechanism for gradual improvement of the agent.

\section{Experimental setup }\label{sec:experiment}

The goals of our experiments are two fold. First, we introduce our RL-agent based on the proposed framework with small number of base negotiators and show that the RL-agent, while generalizing over negotiation scenarios performs on average better than state of the art ANAC winning agents. Secondly, we show the  value  of the reviewer mechanism by evaluating and adding new negotiators and subsequently show improvement of our RL-agent. 
\begin{table}[t]
  \caption{Overview of the ANAC 2013 domains. We considered 18 domains with a pair of utility functions.}
  \label{table:domains}
  \begin{tabular}{rll}\toprule
    \textit{Domain} & \textit{Opposition} & \textit{Outcome Space} \\ \midrule
Acquisition & 0.104 & 384\\
Animal & 0.15 & 1152\\
Camera & 0.076 & 3600\\
Coffee & 0.279 & 112\\
Defensive Charms & 0.193 & 36\\
Dog Choosing & 0.002 & 270\\
Fifty Fifty & 0.498 & 11\\
House Keeping & 0.13 & 384\\
Ice-cream & 0.01 & 720\\
Kitchen & 0.219 & 15625\\
Laptop & 0.076 & 27\\
Lunch & 0.246 & 3840\\
Nice Or Die & 0.177 & 3\\
Outfit & 0.049 & 128\\
planes & 0.606 & 27\\
Smart Phone & 0.022 & 12000\\
Ultimatum & 0.319 & 9\\
Wholesaler & 0.128 & 56700\\
 \bottomrule
  \end{tabular}
\end{table}

We analyzed our proposed system in 18 domains of ANAC 2013 with cardinality of outcome space ranging from 3 to 56700 and opposition~\cite{baarslag2013evaluating} ranging from 0.002 to 0.606 as shown in Table~\ref{table:domains}. All the negotiation experiments are conducted using the NEGotiation MultiAgent System (NegMAS) platform~\cite{Yasser2020negmas, negmas}.  For the purpose of calculating benchmarks and evaluating performance  we used the given preference profiles of ANAC 2013 for each domain. Among the negotiation settings, the reserved value is kept zero and the discount factor is ignored for all negotiations. Moreover, we used min-max normalisation for normalising the utility values between 0 and 1. For performance comparisons average utility values are calculated on negotiation data obtained from 50 to 100 negotiations between a pair of agents for each negotiation scenario.
\par
For training a bidding strategy against a given negotiator we generate a random utility function for the opponent and used fixed self utility function. This ensures the that maximum entropy reinforcement learning algorithm can learn a stochastic policy that performs well in varied negotiation scenarios. Additionally, the training of strategies are done in a single domain and evaluations are done in all 18 domains. For our experiments all strategies are trained on the \emph{camera} domain and tested across other domains. Moreover, for simplicity all hyperparameters of SAC algorithm were kept fixed while training against different negotiators. Training of RL-agent was done using the TF-Agents~\cite{TFAgents} library.
\par
For the opponent classifier we used a window length of 20 and the model consists of three consecutive 1D-CNN layers followed by two Dense layers. The training data for the classifier was generated by multiple simulations of the base negotiators while training the RL strategies against them. 
Additionally, the classifier is trained only on the data generated in the camera domain while the same classifier has been used in the evaluation for all other domains. The training of the classifier was done using TensorFlow library~\cite{tensorflow2015-whitepaper}.\par 
Evaluations of our proposed framework are done by first instantiating an RL-agent with the number of base negotiators $n=1$ and then increasing it to $n=3$ with the help of Reviewer mechanism. We used parameters $\alpha=\beta=1.1$ for the Reviewer and for simplicity, opted for a pure switching mechanism instead of combining strategies. Furthermore, we conducted t-tests with Bonferroni’s conservative multiple-comparisons correction~\cite{bonferroni} for analysing if the difference in mean utilities are statistical significant.

\begin{figure*}[!ht]
\begin{subfigure}{0.25\textwidth}
\includegraphics[width=1\linewidth, height=5cm]{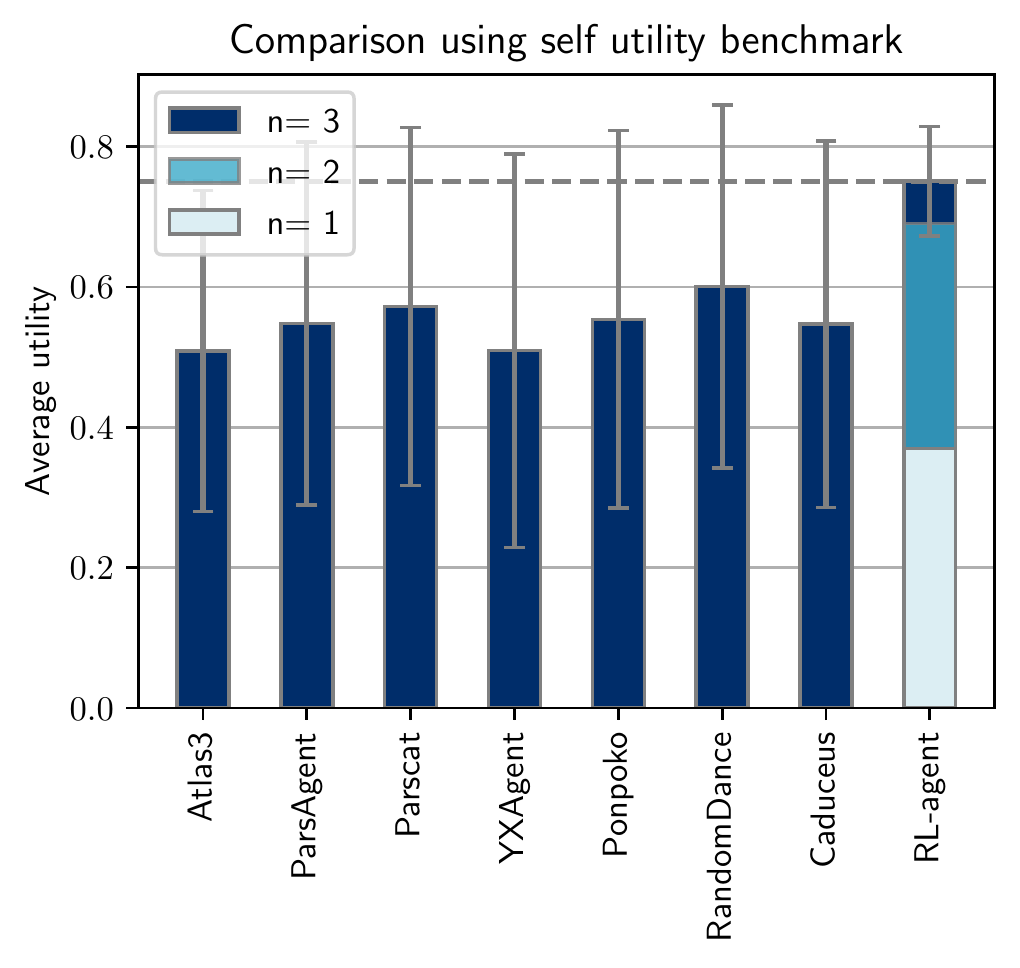} 
\caption{}
  \label{fig:UtilityBenchmarkSmallFinal}
  \Description{Comparison of the average performance of our framework with the average performance of 7 ANAC winning agents. }
\end{subfigure}
\begin{subfigure}{0.25\textwidth}
\includegraphics[width=1\linewidth, height=5cm]{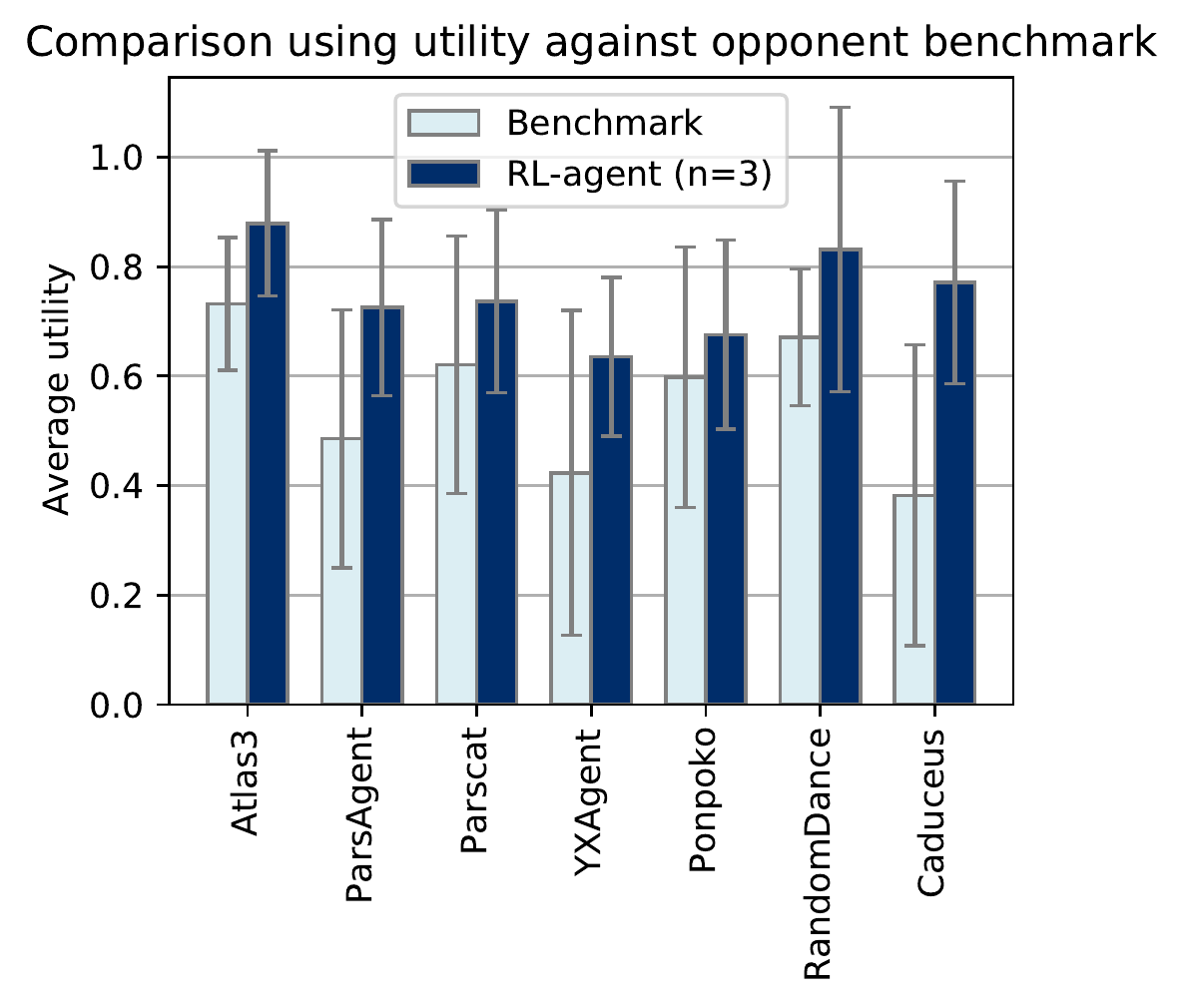}
 \caption{}
  \label{fig:utilityAgainstSmallFinal}
  \Description{Comparison of our framework against opponent benchmark consisting of 7 ANAC winning agents. }
\end{subfigure}
\begin{subfigure}{0.48\textwidth}
\includegraphics[width=1\linewidth, height=5cm]{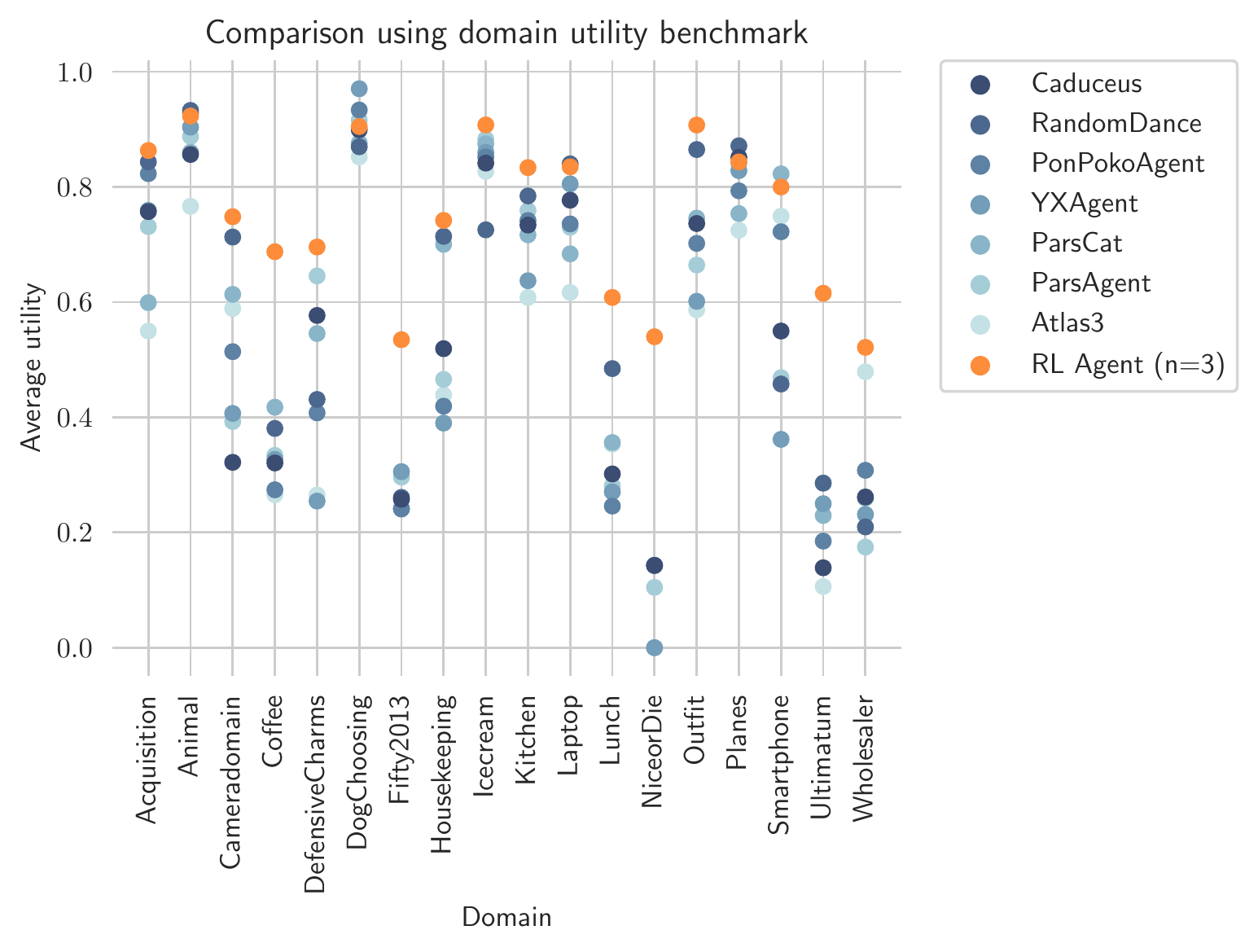}
\caption{}
  \label{fig:UtilityBenchmarkDomainSmallFinal}
  \Description{Comparison of our framework against agent benchmark consisting of 7 ANAC winning agents.}
\end{subfigure}
\caption{(a) Comparison of RL-agent with  self utility benchmark consisting of 7 ANAC winning agents. It also shows the performance of RL-agent with $n=1, n=2$ and $n=3$. (b) Comparison of RL-agent with utility against opponent benchmark  consisting of 7 ANAC winning agents. (c) Comparison of RL-agent with domain benchmark consisting of 18 domains.}
\label{fig:small_benchmark}
\end{figure*}

\section{Results}\label{sec:result}

In this section we present the results according to the experimental setup of Section~\ref{sec:experiment}. First we will present the detailed results of our experiments with 7 ANAC winning agents. Next we present the evaluations of the Reviewer mechanism.
Finally, we present a summarised result against 47 GENIUS agents. For comparison, we created three different benchmarks and then compared the performance of RL-agent against each of the benchmarks. Now, $U_{a}^{a\times b: d}$ denotes the average utility achieved by agent $a$ against $b$ in domain $d$  over 100 runs with two different utility functions. $A$ and $D$ denotes the set of agents and domains respectively over which the benchmark is calculated and $|.|$ denotes the cardinality of a set.
\begin{enumerate}
  \item \emph{Self utility benchmark}: In this benchmark, score of an agent $S_a = \frac{1}{|A|\times|D|}\sum_{d\in D}\sum_{b\in A} U_{a}^{a\times b: d}$ 
  is the mean utility acquired by the agent $a$ when negotiating with every agent $b \in A$ in all negotiation scenarios.
  \item \emph{Utility against opponent benchmark }:  In this benchmark,  score against an agent $a$,  $O_a = \frac{1}{(|A|-1)\times|D|}\sum_{d\in D}\sum_{b\in A/{a}} U_{b}^{a\times b: d}$  
  denotes the mean utility acquired by agents $b \in A/a$ while negotiating with agent $a$ in all negotiation scenarios.
   \item \emph{Domain utility benchmark}: In this benchmark,  score of a domain $d$, $D_d = \frac{1}{|A|}\sum_{b\in A} U_{a}^{a\times b:d}$  
   denotes the mean utility obtained by all agents $a \in A$ in domain $d$, while negotiating with every agent $b \in A$.
\end{enumerate}

\subsection{Performance against ANAC Winning Agents}
Before evaluating our agent, we first calculate the benchmark scores with 7 ANAC winning agents and then compare the score of our RL agent with each benchmarks as shown in Figure~\ref{fig:small_benchmark}. The agents selected are Atlas3, ParsAgent, RandomDance, ParsCat, AgentYX, Caduceus  and PonpokoAgent
~\cite{fujita2017sixth, aydougan2018anac, ANAC2016}\footnote{Atlas3 (2015 winner), ParsAgent (2015 $2^{nd}$ position), RandomDance (2015 $3^{rd}$ position), Caduceus (2016 winner), ParsCat (2016 $2^{nd}$ position), AgentYX (2016 $2^{nd}$ position) and PonpokoAgent (2017 winner)}. 
It is clearly visible from Figure~\ref{fig:small_benchmark}, that our RL-agent outperformed all  7   agents in all the benchmarks. The error bars in Figure~\ref{fig:small_benchmark}a and Figure~\ref{fig:small_benchmark}b denote the standard deviation of average utilities obtained over the domains. In comparison with self utility benchmark, RL-agent performed $25\%$ better than the agent which acquired highest average utility in that benchmark as shown in Figure~\ref{fig:small_benchmark}a. Overall, our agent's score was $37\%$ higher than the average scores of all other agents. Glancing at the error bars, one can understand that the RL-agent has the minimum standard deviation among all other agents and hence shows the robustness of the agent in varied negotiation domains. In comparison with utility against opponent benchmark, the average utility obtained by the RL-agent outperformed the benchmark scores in a range of $11\%$ to $50\%$   as shown in Figure~\ref{fig:small_benchmark}b. This shows that the RL-agent outperforms the average score of the  opponents against each agent.
Proceeding to the comparison with domain benchmark illustrated in Figure~\ref{fig:small_benchmark}c, one can clearly visualize that the scores of RL-agent is better than the highest score by any agent in 13 out of 18 domains. In fact the average score of RL-agent is more than the utility benchmark by a range of $4\%$ to $450\%$.

\begin{figure*}[ht]
  \centering
  \includegraphics[width=1\textwidth, height=0.3\textheight]{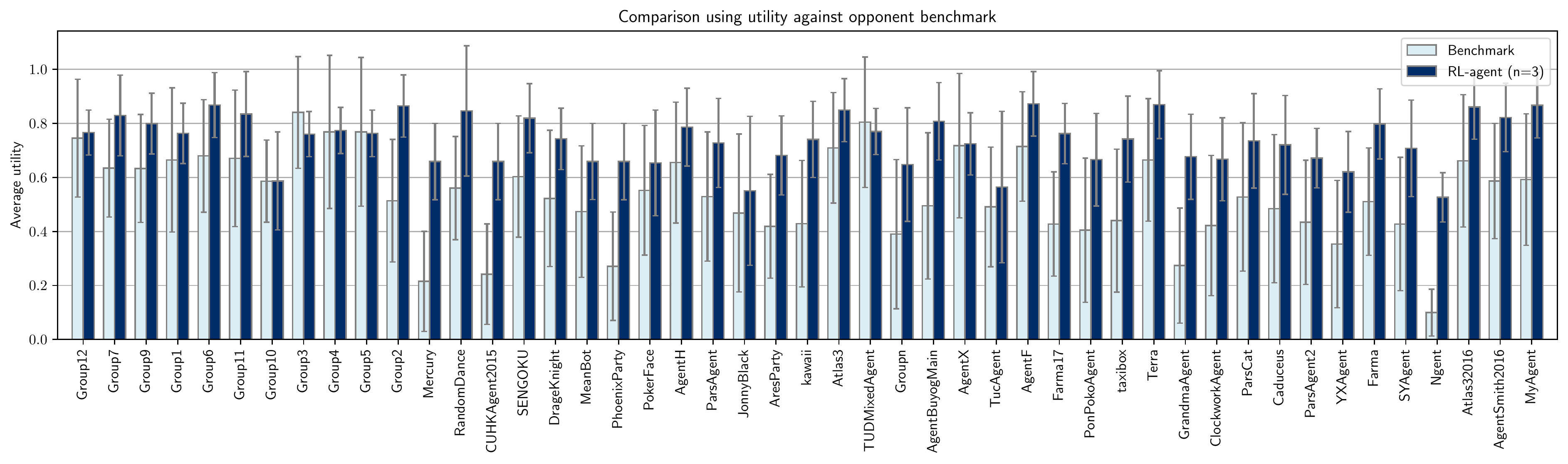}
  \caption{Comparison of RL-agent with  utility against opponent benchmark  consisting of 47 GENIUS Agents in 18 domains.}
  \label{fig:utilityAgainstLargeFinal}
  \Description{Comparison of RL-agent with  utility against opponent benchmark  consisting of 47 GENIUS Agents in 18 domains.}
\begin{subfigure}{0.59\textwidth}
\includegraphics[width=1\linewidth, height=5cm]{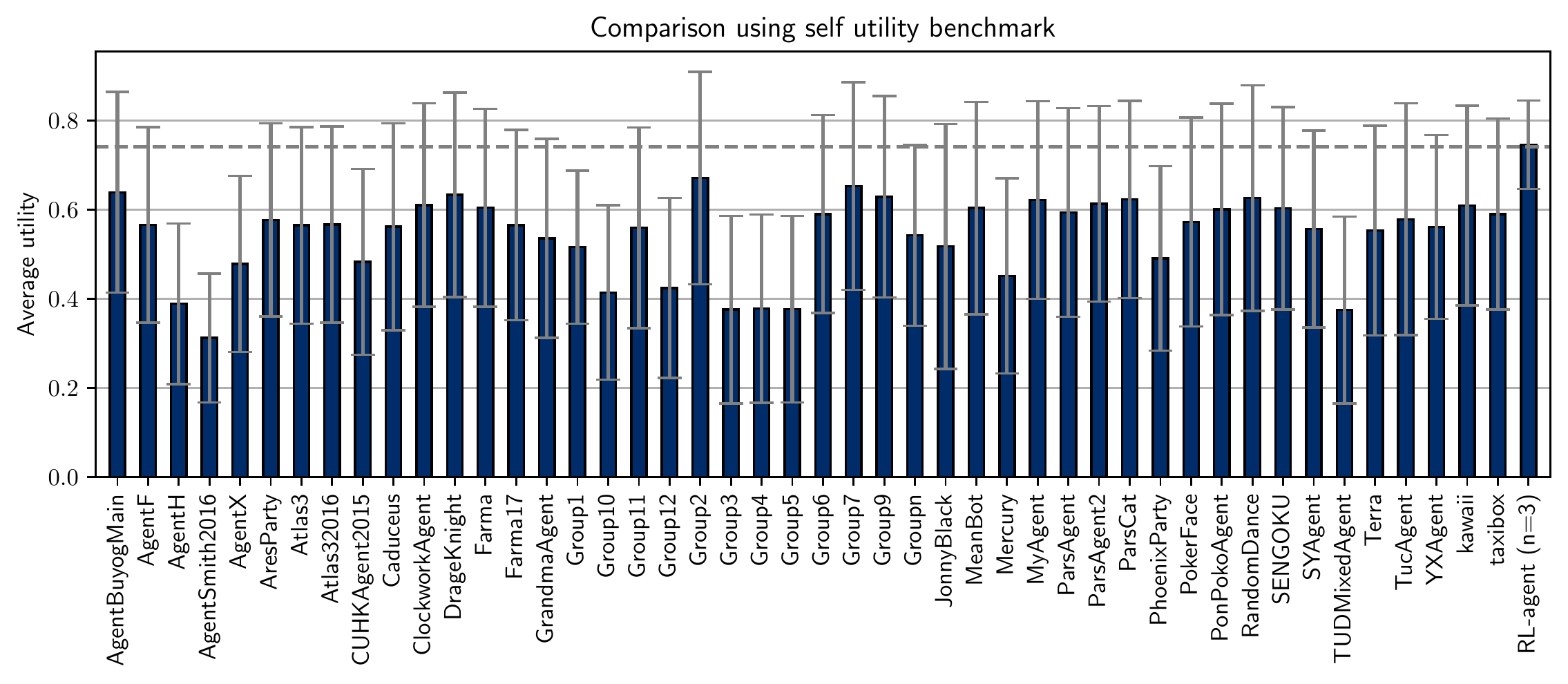} 
\caption{}
  \label{fig:UtilityBenchmarkLargeFinal}
  \Description{Comparison of the performance of RL-agent with  self utility benchmark consisting of 47 GENIUS Agents}
\end{subfigure}
\begin{subfigure}{0.39\textwidth}
\includegraphics[width=1\linewidth, height=5cm]{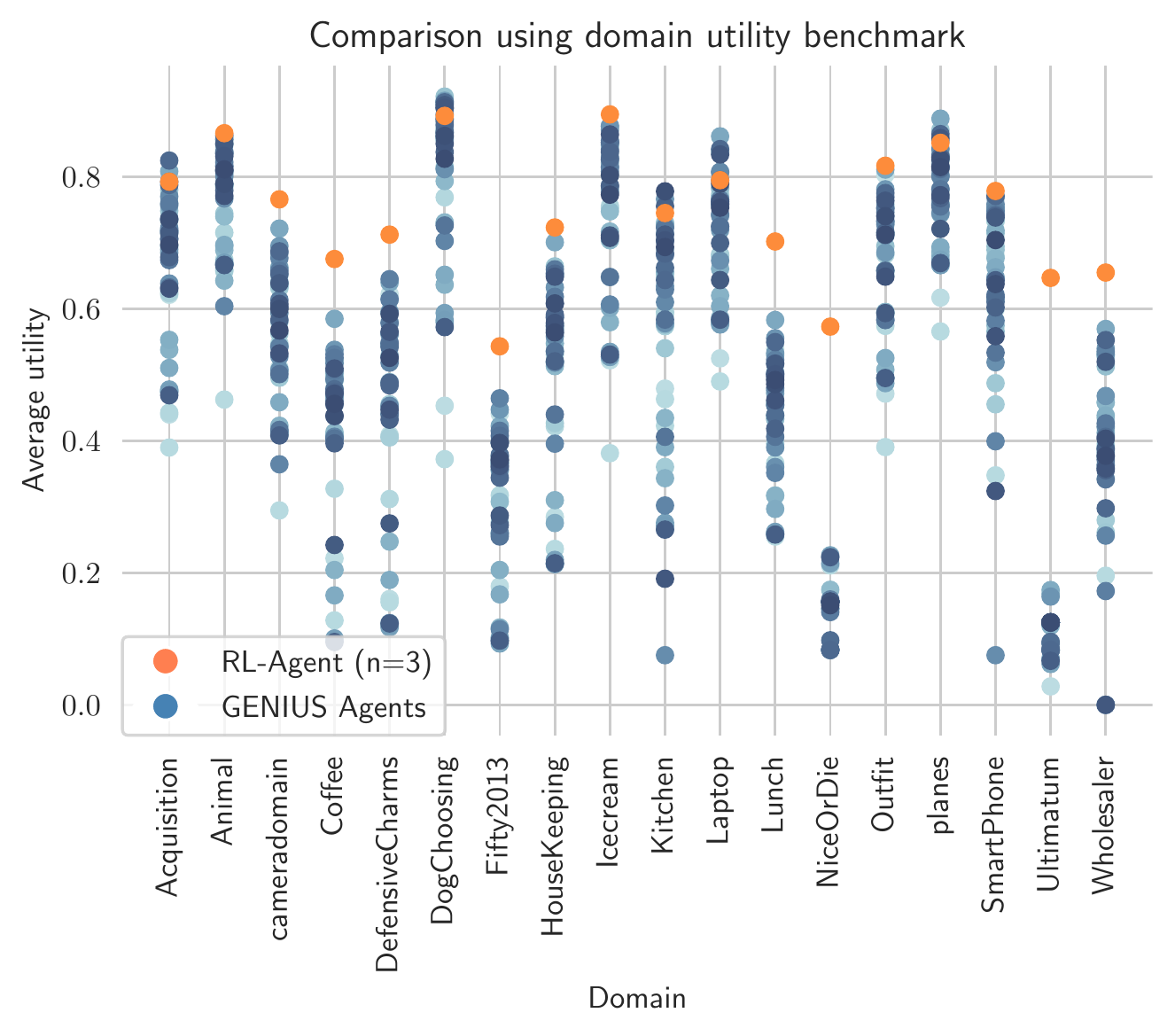}
\caption{}
  \label{fig:UtilityBenchmarkDomainlargeFinal}
  \Description{Comparison of the performance of RL-agent with  domain benchmark  consisting of 18 domains.}
\end{subfigure}
\caption{Comparison of the performance of RL-agent (a) with  self utility benchmark consisting of 47 GENIUS Agents and (b) with  domain benchmark  consisting of 18 domains. }
\label{fig:large_benchmark}
\end{figure*}

\subsection{Performance of Reviewer}
Although the final comparison shown in Figure~\ref{fig:small_benchmark} is the instance of RL-agent with three negotiator-strategy pairs, the framework was first initialized with only one negotiator, that is, with the Random negotiator. One RL based strategy was trained against it and included in the negotiator-strategy pair. Opponent classifier training was not needed as number of base negotiator $n=1$. As expected the performance of the initial RL-agent was poor as shown in Figure~\ref{fig:UtilityBenchmarkSmallFinal}. Next we introduced a simple time dependent agent called Boulware agent~\cite{faratin1998negotiation}, to the Reviewer. After receiving acceptance from reviewer, new RL based strategy was added to the strategy pool and the opponent classifier was trained for $n=2$. In a similar way, a new behavioural strategy based agent, Naive tit-fot-tat and corresponding RL based strategy was included in the RL-agent and classifier was trained for $n=3$. Furthermore, the strategy trained against random negotiator was replaced by the RL strategy trained against Naive tit-for-tat as per the evaluation of the Reviewer. Moving ahead, we introduced other agents from the pool of 7 ANAC agents to the Reviewer, but the Reviewer rejected the inclusion of any additional agents. It is to be noted that all evaluations by the Reviewer were also restricted to a single domain only (camera domain). The performance of the RL-agent with different number of base negotiators is shown in Figure~\ref{fig:small_benchmark}a. It can be clearly seen that the performance of the RL-agent has increased with the addition of  strategies and hence shows the significance of the Reviewer mechanism in our framework. To keep the visualisations simple, further comparisons with other benchmarks as shown in Figure~\ref{fig:small_benchmark}b and Figure~\ref{fig:small_benchmark}c are only shown with number of base negotiators $n=3$.
The results obtained have two fold significance, firstly it shows that the bidding behaviour of all 7 opponent agents can be approximated by a piece-wise function of Boulware type, behavioural type or random type negotiator. So the strategy switching technique within a negotiation session,  with strategies specifically trained for three basic negotiators works undoubtedly well against these 7 opponents. Secondly, it shows the Reviewer mechanism plays a key role in addition of base strategies and thereby improving the RL-agent's overall performance.\par
\subsection{Performance against GENIUS Agents}
To show the versatility of the proposed framework, we chose the already created RL-agent with only 3 base negotiators and evaluated it against the benchmarks scores of 47 GENIUS agents which includes ANAC competitors from year 2015 to 2017. The results are illustrated in Figure~\ref{fig:large_benchmark} and Figure~\ref{fig:utilityAgainstLargeFinal} where the error bars denote the standard deviation over the domains.
In comparison with self utility  benchmark, our RL-agent performed  better than other agents in the range of $11\%$ to $139\%$. The average improvement is $37.4\%$ with an improvement of more than $50\%$ against 12 agents as shown in Figure~\ref{fig:UtilityBenchmarkLargeFinal}. The relatively low standard deviation marks the robustness of our agent in different domains. Next, comparison with the utility against opponent benchmark  as shown in Figure~\ref{fig:utilityAgainstLargeFinal}, reveals that the RL-agent outperformed in 40 out of 47 comparisons. Individual performance improvement ranges from $1\%$ to $428\%$ with an improvement of over $50\%$ against 19 agents and $100\%$ against 5 agents. Finally, comparison with domain benchmark  is shown in Figure~\ref{fig:UtilityBenchmarkDomainlargeFinal}, where it is visible that the scores of RL-agent is better than the highest score by any agent in 13 out of 18 domains. Additionally, the RL-agent outperformed the average  benchmark scores in all the domains. The performance improvement  over the  benchmark scores in each domain ranges from $10\%$ to $488\%$ with an improvement of at least $50\%$ in 7 domains. Finally, we calculate the statistical significance of the differences. For the utility differences to be statistically significant, following Bonferroni’s conservative multiple-comparisons correction  the p-values of each t-tests in self utility  benchmark, utility against opponent benchmark  and domain benchmark should be less than 0.0011, 0.0011 and 0.0028 respectively.  It turns out that for self utility benchmark, differences of utility in 30 out of 47 comparisons were statistically significant whereas in utility against opponent benchmark  32 out of 47 were statistically significant. In case of domain benchmark, all differences of utilities were statistically significant.

\section{Conclusion and Future Work}\label{sec:conclusion}
In this work we proposed an autonomous negotiating agent framework with four components: negotiators paired with trained strategies, an opponent classifier, a switching mechanism and a reviewer mechanism. Strategies included are RL based strategies whereas strategy switching depends on  the classification probabilities of the opponent classifier. The proposed opponent classifier classifies the opponent's bidding behaviour with respect to the base negotiator's behaviour at every time step thus allows the agent to switch or combine strategies within a single negotiation session. These functionalities together gives our RL-agent versatility even with a small pool of base negotiators as illustrated in our evaluations.  Furthermore, the reviewer mechanism helps in the decision making of adding more negotiators and strategies to the existing pool of base negotiators and strategies. This helps in incremental improvement of the RL-agent and also restricts unnecessary addition of base entities.\par 
In our experimental setup, all training and evaluations were done by removing the discount factor and keeping the reserved utility value as zero. It would be interesting to obtain and compare the results with varying discount factors and reserved utility values. Moreover, while training bidding strategies, we have used the same acceptance strategies against all negotiators. At the same time, it has been noticed that the performance of trained strategy depends on the choice of acceptance strategy. So our future work involves training the acceptance strategy together with the bidding strategy. Also, to show the concept of reviewer mechanism, we have implemented an evaluation based reviewer. Another interesting direction could be using  unsupervised clustering algorithms on the negotiators bidding behaviour to differentiate a new negotiator from the pool of negotiators. That will remove the additional step of training a new strategy each time for evaluation by the reviewer and at the same time will give us a concrete picture about the type of base negotiators that make the RL-agent perform better across various negotiation scenarios and against varied opponents.








\bibliographystyle{ACM-Reference-Format}
\bibliography{main}


\end{document}



\pagestyle{fancy}
\fancyhead{}


\maketitle 


\section{Experimental setup}
As part of supplementary material we are providing additional information to make our work reproducible. In our proposed method we create several policies which are trained with reinforcement learning algorithms. Additionally, we also train a classifier with 1D-CNN based deep learning technique that is capable of classifying an unknown opponent and enables the agent to switch mechanism within a single negotiation round. In this material firstly, we will provide the detailed implementation and the corresponding hyperparameters for the Soft-Actor-Critic (SAC) algorithm~\cite{haarnoja2018soft, haarnoja2018soft2} used for training bidding strategies against the base negotiators. Secondly we layout the model architecture and the corresponding hyperparameters for the 1D-CNN based opponent classifier.\par
The hyperparameters of SAC algorithm used for our experiments while training strategies are provided in Table~\ref{tbl:sac}. We used TF-Agents~\cite{TFAgents} library for the implementation of the SAC algorithm. The first version of SAC~\cite{haarnoja2018soft} uses a fixed entropy temperature $\alpha$. Though the performance of the original SAC was quite impressive, $\alpha$ turned out to be a very sensitive hyperparameter. To remedy this, in the second version of SAC~\cite{haarnoja2018soft2} alpha is converted to a trainable parameter and we have used this version of SAC algorithm. Furthermore, It is to be noted that in SAC, two critic networks are created with the same structure each one with its own layers and weights. Second one is usually known as target critic network. After every \emph{target update period} train steps, the weights from critic network are copied with smoothing via \emph{target update $\tau$} to target critic network. Additionally,   NEGotiation MultiAgentSystem (NegMAS)~\cite{Yasser2020negmas, negmas} platform is used for the purpose of creating the negotiation environment on which the reinforcement learning algorithms were trained. All other negotiation simulations for benchmarking and results are also done using the same platform.\par
\begin{figure}[ht]
  \centering
  \includegraphics[width=0.25\textwidth]{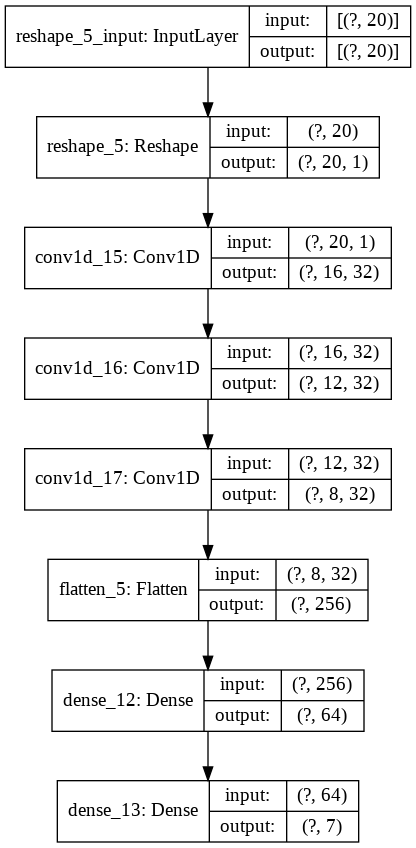}
  \caption{Classifier model having three consecutive 1D-CNN layers followed by two Dense layer for $k=20$ and $N=7$}
  \label{fig:classifier_model}
  \Description{Classifier model having three consecutive 1D-CNN layers followed by two Dense layer for $k=20$ and $n=7$}
\end{figure}

The opponent classifier as proposed in our work, is a 1D-CNN based classifier and is trained using TensorFlow library~\cite{tensorflow2015-whitepaper}. The model has three consecutive 1D-CNN layers, followed by a two fully connected layers as shown in Figure~\ref{fig:classifier_model}. The parameter $k$ denotes the window length or the length of the input sequence to the classifier and the parameter $n$ denotes the number of classes or base negotiators. We have separately tested the classifier with base negotiators ranging from 2 to 7. For our final results we only required the model with 2 and 3 classes.
Here Figure~\ref{fig:classifier_model} illustrates a model with $k=20$ and $n=7$, that is, input consisting of 20 opponent offers  and outputs a set of probabilities for each of the seven classes. The hyperparameters of the opponent classifier are given in Table~\ref{tbl:1dcnn}.

\begin{table}[ht]
  \caption{Hyperparameters for opponent classifier}
    \begin{tabular}{rl}\toprule \label{tbl:1dcnn}
    \textit{Hyperparameter} & \textit{values}\\ \midrule
 Training epochs & 30-40\\
 1D-CNN layer activation function & relu\\
 1D-CNN layer filter size & 32\\
 1D-CNN kernel size & 5\\
Dense layer activation function & relu\\
Final layer activation function & softmax\\
Learning rate & 0.0001\\
Batch size & 50\\
Loss function & Categorical Crossentropy\\
Optimizer & Adam Optimizer\\\bottomrule
  \end{tabular}
\end{table}

\begin{table*}[ht] 
  \caption{Hyperparameters for SAC algorithm}
  \label{tab:locations}
  \begin{tabular}{rll}\toprule \label{tbl:sac}
    \textit{Hyperparameters} & \textit{values} & \textit{Description}   \\ \midrule
    Epochs & 7000-20000 & Number of iterations to run and train agent. \\
    Initial collect steps & 500 & Number of steps for uniform-random action selection, before running real policy. \\
    Replay buffer capacity & 1000000 &  Maximum length of replay buffer.\\
    Batch size & 128 & Minibatch size for SGD. \\ 
    Critic fc layers & (256,512) & fully connected parameters, where each item is the number of units in the layer.\\
    Critic activation fn & relu & Activation function for critic network. \\
    Critic learning rate & 3e-2 & Learning rate for critic network. \\
    Critic Optimizer & Adam Optimizer & The optimizer used for the critic network.\\
    Critic loss weight & 0.5 & The weight on critic loss.\\
    Actor fc layers & (256, 512, 512) & fully connected parameters, where each item is the number of units in the layer.\\
    Actor activation fn & relu & Activation function for actor network. \\
    Actor learning rate & 3e-5 & Learning rate for actor network.\\
    Actor Optimizer & Adam Optimizer & The  optimizer used for the actor network.\\
    Actor loss weight & 1 & The weight on actor loss.\\
    $\alpha$ learning rate & 3e-3 & Learning rate for $\alpha$ network.\\
    $\alpha$ Optimizer & Adam Optimizer & The  optimizer used for the $\alpha$ network.\\
    $\alpha$ loss weight & 1 & The weight on $\alpha$ loss.\\
    Target update $\tau$ & 0.005 & Factor for soft update of the target networks. \\
    Target update period & 1 & Period for soft update of the target networks.\\
    TD error loss function & mean squared error & A function for computing the element wise TD errors loss. \\
    $\gamma$  & 0.99 &  Discount factor.\\
    Reward scale factor & 1 & Multiplicative scale for the reward.\\
    Initial log alpha & 0.0 & Initial value for log alpha.\\    \bottomrule
  \end{tabular}
\end{table*}




\bibliographystyle{ACM-Reference-Format} 
\bibliography{supp}
